\documentclass[journal,onecolumn,draftclsnofoot]{IEEEtran}
\usepackage{graphicx}
\usepackage{amsfonts}
\usepackage{amssymb}
\usepackage{amsmath}
\usepackage{mathdots}
\usepackage{mathtools}
\usepackage{amsthm}
\usepackage{subfigure}

\usepackage{mathrsfs}

\usepackage{enumitem}
\usepackage{url}
\usepackage{dsfont}

\usepackage{multirow,tabularx}
\usepackage{flushend}
\usepackage[numbers,sort&compress]{natbib}

\usepackage{subfig}
\usepackage{epstopdf}

\usepackage{algorithm}
\usepackage{algorithmic}

\usepackage{wrapfig}

\usepackage{color}

\usepackage{tablefootnote}

\newtheorem{theorem}{Theorem}
\newtheorem{lemma}{Lemma}

\newtheorem{definition}{Definition}
\newtheorem{corollary}{Corollary}
\newtheorem{proposition}{Proposition}
\newtheorem{open problem}{Open Problem}

\newtheorem{innercustomthm}{Theorem}
\newenvironment{customthm}[1]
{\renewcommand\theinnercustomthm{#1}\innercustomthm}
{\endinnercustomthm}
\newtheorem{innercustomlemma}{Lemma}
\newenvironment{customlemma}[1]
{\renewcommand\theinnercustomlemma{#1}\innercustomlemma}
{\endinnercustomlemma}

\begin{document}
\bstctlcite{IEEEexample:BSTcontrol}
\title{Fast Multi-label Learning* \footnote{*The original version of this paper has been published on IJCAI 2021.}}

\author{Xiuwen Gong,
        Dong Yuan,
        Wei Bao
\thanks{X. Gong, D. Yuan, W. Bao are with the Faculty of Engineering,
The University of Sydney, Australia (E-mails: xiuwen.gong@sydney.edu.au; dong.yuan@sydney.edu.au; wei.bao@sydney.edu.au)}
}

\maketitle

\begin{abstract}
Embedding approaches have become one of the most pervasive techniques for multi-label classification. However, the training process of embedding methods usually involves a complex quadratic or semidefinite programming problem, or the model may even involve an NP-hard problem. Thus, such methods are prohibitive on large-scale applications. More importantly, much of the literature has already shown that the binary relevance (BR) method is usually good enough for some applications. Unfortunately, BR runs slowly due to its linear dependence on the size of the input data.  The goal of this paper is to provide a simple method, yet with provable guarantees, which can achieve competitive performance without a complex training process. To achieve our goal, we provide a simple stochastic sketch strategy for multi-label classification and present theoretical results from both algorithmic and statistical learning perspectives. Our comprehensive empirical studies corroborate our theoretical findings and demonstrate the superiority of the proposed methods.
\end{abstract}

\begin{IEEEkeywords}
Online Classification, Multi-label, Metric Learning, $k$-Nearest Neighbour ($k$NN).
\end{IEEEkeywords}

\IEEEpeerreviewmaketitle

\section{Introduction}
\label{Introduction}

Multi-label classification \cite{Prabhu14,DBLP:conf/icml/YenHRZD16,DBLP:journals/pami/LiuXTZ19,DBLP:conf/aaai/GongYB20}, in which each instance can belong to multiple labels simultaneously, has significantly attracted the attention of researchers as a result of its wide range of applications, which range
from document classification and automatic image annotation to video annotation.
For example, in automatic image annotation, one needs to automatically predict relevant keywords, such as  \textit{beach}, \textit{sky} and \textit{tree}, to describe a natural scene image. When classifying documents, one may need to classify them into different groups, such as \textit{Science}, \textit{Finance} and \textit{Sports}. In video annotation, labels such as \textit{Government}, \textit{Policy} and \textit{Election} may be needed to describe the subject of the video.

A popular strategy in multi-label learning is binary relevance (BR)\cite{Tsoumakas2010}, which independently trains a linear regression model for each label independently. Recently, some sophisticated models are developed to improve the performance of BR. For example, embedding approaches \cite{conf/nips/HsuKLZ09,DBLP:conf/nips/ChenL12,DBLP:conf/icml/Yu0KD14,DBLP:journals/jmlr/LiuT17,DBLP:journals/jmlr/LiuTM17} have become popular techniques. Even though embedding methods improve the prediction performance of BR to some extent, their training process usually involves a complex quadratic or semidefinite programming problem, as in \cite{conf/icml/ZhangS12}, or their model may involve an NP-hard problem, as in \cite{DBLP:conf/icml/Yu0KD14} and \cite{Bhatia2015}. Thus, these kinds of methods are prohibitive on large-scale applications. Much of the literature, such as \cite{DBLP:journals/pai/LuacesDBCB12}, \cite{DBLP:journals/pr/MadjarovKGD12} and \cite{BinaryrelevanceBRmethodclassifier}, has already shown that BR with appropriate base learner is usually good enough for some applications, such as document classification \cite{BinaryrelevanceBRmethodclassifier}. Unfortunately, BR runs slowly due to its linear dependence on the size of the input data.
The question is how to overcome these computational obstacles yet obtain comparable results with BR.

To address the above problem, we provide a simple stochastic sketch strategy for multi-label classification. In particular, we carefully construct a small sketch of the full data set, and then use that sketch as a surrogate to perform fast optimization. This paper first introduces stochastic $\sigma$-subgaussian sketch, and then proposes the construction of a sketch matrix based on Walsh-Hadamard matrix to reduce the expensive matrix multiplications of $\sigma$-subgaussian sketch. From an algorithmic perspective, we provide provable guarantees that our proposed methods are approximately as good as the exact solution of BR. From a statistical learning perspective, we provide the generalization error bound of multi-label classification using our proposed stochastic sketch model.

Experiments on various real-world data sets demonstrate the superiority of the proposed methods. The results verify our theoretical findings. We organize this paper as follows. The second section introduces our proposed stochastic sketch for multi-label classification. The third section provides the provable guarantees for our algorithm from both algorithmic and statistical learning perspectives, and experimental results are presented in the fourth section. The last section provides our conclusions.


\section{Stochastic Sketch for Multi-label Classification}
\label{SS}

Assume that $x^{(i)} \in \mathbb{R}^{p \times 1} $ is a real vector representing an input (instance), and $y^{(i)} \in \{0,1\}^{q \times 1}$ is a real vector representing the corresponding output $(i \in \{1\ldots n\})$. $n$ denotes the number of training samples. The input matrix is $X \in \mathbb{R}^{n\times p} $ and the output matrix is $Y \in \{0,1\}^{n\times q} $. $\langle \cdot ,\cdot \rangle$ and $\mathbf{I}_{n\times n}$ represent the inner product and the $n\times n$ identity matrix, respectively. We denote the transpose of the vector/matrix by the superscript $'$ and the logarithms to base 2 by $log$. Let $||\cdot ||_2$ and $||\cdot ||_F$ represent the $l_2$ norm and Frobenius norm, respectively. Let $V \in \mathbb{R}^{p\times q}$ be the regressors and $N(0,1)$ denote the standard Gaussian distribution.

A simple linear regression model for BR \cite{Tsoumakas2010} learns the matrix $V$ through the following formulation:
\begin{equation}\label{2}
	\min_{V\in \mathbb{R}^{p\times q}}  \frac{1}{2}||XV - Y||^2_F
\end{equation}
Assuming that $n>p$ and $n>q$, the computational complexity for this problem is $\mathcal{O}(npq+np^2)$ \cite{DBLP:books/daglib/0086372}.
The computational cost of an exact solution for problem \ref{2} will be prohibitive on large-scale settings.
To solve this problem, we construct a small sketch of the full data set by stochastic projection methods, and then use that sketch as a surrogate to perform fast optimization for problem \ref{2}. Specifically, we define a sketch matrix $S \in \mathbb{R}^{m \times n}$ and $S \neq 0$, where $m < n$ is the projection dimension and $0$ is the zero matrix with all the zero entries. The input matrix $X$ and output matrix $Y$ are approximated by their sketched matrix $SX$ and $SY$, respectively. We aim to solve the following sketched problem of problem \ref{2}.
\begin{equation}\label{s2}
	\min_{V\in \mathbb{R}^{p\times q}}  \frac{1}{2}||SXV - SY||^2_F
\end{equation}
Motivated by \cite{LMNN,journals/ftml/Kulis13,Bhatia2015}, we use a $k$-nearest neighbor ($k$NN) classifier in the embedding space for prediction, instead of using an expensive decoding process \cite{conf/icml/ZhangS12}. Next, we introduce two kinds of stochastic sketch methods.

\subsection{Stochastic $\sigma$-Subgaussian Sketch}

The entries of a sketch matrix can be simply defined as i.i.d random variables from certain distributions, such as Gaussian distribution and Bernoulli distribution. \cite{DBLP:journals/rsa/Matousek08} has already shown that each of these distributions is a special case of Subgaussian distribution, which is defined as follows:

\begin{definition}[$\sigma$-Subgaussian] \label{ssdef1}
	A row $s_i \in \mathbb{R}^{n} $ of the sketch matrix $S$ is $\sigma$-Subgaussian, if it has zero mean and for any vector $\zeta \in \mathbb{R}^{n} $ and $\epsilon \geq 0$, we have
	\begin{equation*}\label{defeq1}
		\begin{split}
			P(|\langle s_i,\zeta \rangle|\geq \epsilon ||\zeta||_2 ) \leq 2 e^{-\frac{n \epsilon^2}{2\sigma^2}}
		\end{split}
	\end{equation*}
\end{definition}
Clearly, a vector with i.i.d standard Gaussian entries or Bernoulli entries is 1-Subgaussian. We refer any matrix $S \in \mathbb{R}^{m \times n}$ to a Subgaussian sketch if its rows are zero mean, 1-Subgaussian, and
with the covariance matrix $cov(s_i)=\mathbf{I}_{n\times n}$.
A Subgaussian sketch is straightforward to construct. However, given the Subgaussian sketch $S \in \mathbb{R}^{m \times n}$, the cost of computing $SX$ and $SY$ is $\mathcal{O}(npm)$ and $\mathcal{O}(nqm)$, respectively. Next, we introduce the following technique to reduce this time complexity.

\subsection{Stochastic Walsh-Hadamard Sketch}

Inspired by \cite{DBLP:journals/siamcomp/AilonC09}, we propose to construct the sketch matrix based on Walsh-Hadamard matrix to reduce the expensive matrix multiplications of Subgaussian sketch. Formally, a stochastic Walsh-Hadamard sketch matrix $S \in \mathbb{R}^{m \times n}$ is obtained with i.i.d. rows of the form:
\begin{equation*}\label{WHSketch}
	\begin{split}
		s_i=\sqrt{n}e_iHR, \quad i=1,\cdots,m
	\end{split}
\end{equation*}
where $\{e_1,\cdots,e_m\}$ is a random subset of $m$ rows uniformly sampled from $\mathbf{I}_{n\times n}$, $R \in \mathbb{R}^{n\times n}$ is a random diagonal matrix whose entries are i.i.d. Rademacher variables and $H  \in \mathbb{R}^{n\times n}$ constitutes a Walsh-Hadamard matrix defined as:
\begin{equation*}\label{WHMatrix}
	\begin{split}
		H_{ij}=(-1)^{\langle\mathbb{B}(i)-1,\mathbb{B}(j)-1 \rangle},  \quad i,j=1,\cdots,n
	\end{split}
\end{equation*}
where $\mathbb{B}(i)$ and $\mathbb{B}(j)$ represent the binary expression with $\tau$-bit of $i$ and $j$ (assume $2^{\tau}=n$).

Then, we can employ fast Walsh-Hadamard transform \cite{DBLP:journals/tc/FinoA76} to perform $SX$ and $SY$ in $\mathcal{O}(np \log m )$ and $\mathcal{O}(nq \log m )$.

\section{Main Results}
\label{MainResults}

Since we address problem \ref{s2} rather than directly solving problem \ref{2}, which has great advantages for fast optimization, it is interesting to ask the question: what is the relationship between problem \ref{s2} and problem \ref{2}?  Let $V^*$ and $\hat{V}$ be the optimal solutions of problem \ref{2} and problem \ref{s2}. We define $f(V^*)=||XV^* - Y||^2_F$ and $g(\hat{V})=||SX\hat{V} - SY||^2_F$. We will prove that we can choose an appropriate $m$ such that the two optimal objectives $f(V^*)$ and $g(\hat{V})$ are approximately the same. This means that we can speed up the computation of problem \ref{2}, without sacrificing too much accuracy. Furthermore, we provide the generalization error bound of the multi-label classification problem using our proposed stochastic sketch model. To measure the quality of approximation, we first define the $\delta$-optimality approximation as follows:
\begin{definition}[$\delta$-Optimality Approximation] \label{deltaoptimalsketch}
	Given $\delta \in(0,1)$, $\hat{V}$ is a $\delta$-optimality approximation solution, if
	\begin{equation*}\label{def1deltaoptimalsketch}
		\begin{split}
			(1- \delta) f(V^*) \leq g(\hat{V}) \leq (1+\delta) f(V^*)
		\end{split}
	\end{equation*}
\end{definition}
According to the properties of Matrix norm, we have $g(\hat{V}) \leq ||S||_F f(\hat{V})$, so $g(\hat{V})$ is proportional to $f(\hat{V})$. Therefore, the closeness of $g(\hat{V})$ and $f(V^*)$ implies the closeness of $f(\hat{V})$ and $f(V^*)$.

\subsection{$\sigma$-Subgaussian Sketch Guarantee}

We first introduce the tangent cone, which is used by \cite{VariationalAnalysis}:
\begin{definition}[Tangent Cone] \label{ssdef2}
	Given a set $\mathcal {C}\subseteq \mathbb{R}^{p}$ and $x^* \in \mathcal {C}$, the tangent cone of $\mathcal {C}$ at $x^*$ is defined as
	$\mathcal {K} = clconv \{r \in \mathbb{R}^{p} | r= t(x - x^*)$ for some $t \geq 0$ and $x \in \mathcal {C}\}$,
	where clconv denotes the closed convex hull.
\end{definition}
The tangent cone arises naturally in the convex optimality conditions: any $r \in \mathcal {K}$ defines a feasible direction at
the optimal $x^* $, and optimality means that it is impossible to decrease the objective function by moving in directions belonging to the tangent cone. Then, we introduce the Gaussian width, which is an important complexity measure used by \cite{Gor85}:
\begin{definition}[Gaussian Width] \label{ssdef3}
	Given a closed set $ \mathcal {Y} \subseteq \mathbb{R}^{n}$, the Gaussian width of $\mathcal {Y}$, denoted by $\omega(\mathcal {Y})$, is defined as:
	\begin{equation*}\label{def3gw}
		\begin{split}
			\omega(\mathcal {Y}) = \mathbb{E}_g[\sup_{z \in \mathcal {Y}}|\langle g,z \rangle| ]
		\end{split}
	\end{equation*}
	where $g \thicksim N(0,\mathbf{I}_{n\times n}) $.
\end{definition}
This complexity measure plays an important role in learning theory and statistics \cite{RademacherProcessesandBounding}. Let $\mathbb{S}^{n-1}=\{z\in \mathbb{R}^{n}| ||z||_2=1\}$ be the Euclidean sphere. $X\mathcal {K}$ represents the linearly transformed cone: $\{Xr \in \mathbb{R}^{n} |r \in \mathcal {K}\}$, and we use Gaussian width to measure the width of the intersection of $X\mathcal {K}$ and $\mathbb{S}^{n-1}$. This paper defines $\mathcal {Y}= X\mathcal {K} \cap \mathbb{S}^{n-1}$.  We state the following theorem for guaranteeing the $\sigma$-Subgaussian sketch:

\begin{theorem}\label{subgaussiansketchtheorem2}
	Let $S \in \mathbb{R}^{m \times n}$ be a stochastic $\sigma$-Subgaussian sketch matrix, $c_1$ and $c_2$ be universal constants. Given any $\delta \in(0,1)$ and $m =\mathcal{O}( (\frac{c_1}{\delta})^2 \omega^2(\mathcal {Y})) $,
	then with probability at least $1- 6qe^{-\frac{c_2 m \delta^2 }{\sigma^4}}$, $\hat{V}$ is a $\delta$-optimality approximation solution.
\end{theorem}
The proof sketch of this theorem can be found in the supplementary material.

\textbf{Remark.} Theorem \ref{subgaussiansketchtheorem2} guarantees that the stochastic $\sigma$-Subgaussian sketch method is able to construct a small sketch of the full data set for the fast optimization of problem \ref{2}, while preserving the $\delta$-optimality of the solution.

\begin{table*}
	\caption{The results of Hamming Loss on the various data sets.}
	\label{Hamres}
	\begin{center}
		\begin{footnotesize}
			\begin{sc}
				\begin{tabular}{lcccccccccr}
					\hline
					&                &                 &             &            &\multicolumn{3}{c}{SS+GAU}             &          \multicolumn{3}{c}{SS+WH}    \\
					\!\!\!Data Set\!\!\!            &   \!\!\!BR+LIB\!\!\!       &  \!\!\!BR+$k$NN\!\!\!       & \!\!\!FastXML\!\!\!     & \!\!\!SLEEC\!\!\!      & \!\!\!$m=256$\!\!\! & \!\!\!$m=512$\!\!\!  & \!\!\!$m=1024$\!\!\!         &   \!\!\!$m=256$\!\!\!    &    \!\!\!$m=512$\!\!\!   &  \!\!\!$m=1024$\!\!\! \\
					\hline
					\!\!\!corel5k\!\!\!             &  \!\!\!0.0098\!\!\!        & \!\!\!0.0095\!\!\!         & \!\!\!0.0093\!\!\!       & \!\!\!0.0094\!\!\!     & \!\!\!0.0095\!\!\!   & \!\!\!0.0095\!\!\!   & \!\!\!0.0094\!\!\!          & \!\!\!0.0103\!\!\!           &  \!\!\!0.0102\!\!\!   & \!\!\!0.0099\!\!\!\\
					\!\!\!nus(vlad)\!\!\!           &  \!\!\!0.0211\!\!\!        & \!\!\!0.0213\!\!\!         & \!\!\!0.0209\!\!\!       & \!\!\!0.0207\!\!\!     & \!\!\!0.0221\!\!\!   & \!\!\!0.0218\!\!\!   & \!\!\!0.0216\!\!\!          & \!\!\!0.0230\!\!\!           &  \!\!\!0.0225\!\!\!   & \!\!\!0.0218\!\!\! \\
					\!\!\!nus(bow)\!\!\!            &  \!\!\!0.0215\!\!\!        & \!\!\!0.0220\!\!\!         & \!\!\!0.0216\!\!\!       & \!\!\!0.0213\!\!\!     & \!\!\!0.0227\!\!\!   & \!\!\!0.0223\!\!\!   & \!\!\!0.0222\!\!\!          & \!\!\!0.0229\!\!\!           &  \!\!\!0.0226\!\!\!   & \!\!\!0.0223\!\!\!\\
					\!\!\!rcv1x\!\!\!               &  \!\!\!0.0017\!\!\!        & \!\!\!0.0019\!\!\!         & \!\!\!0.0019\!\!\!       & \!\!\!0.0018\!\!\!     & \!\!\!0.00189\!\!\!  & \!\!\!0.00188\!\!\!  & \!\!\!0.00187\!\!\!         & \!\!\!0.00199\!\!\!          &  \!\!\!0.00195\!\!\!  & \!\!\!0.00192\!\!\!\\
					\hline
				\end{tabular}
			\end{sc}
		\end{footnotesize}
	\end{center}
\end{table*}

\begin{table*}
	\caption{The results of Example-F1 on the various data sets.}
	\label{ExampleF1}
	\begin{center}
		\begin{footnotesize}
			\begin{sc}
				\begin{tabular}{lcccccccccr}
					\hline
					&                &                 &             &            &\multicolumn{3}{c}{SS+GAU}             &          \multicolumn{3}{c}{SS+WH}    \\
					\!\!\!Data Set\!\!\!            &   \!\!\!BR+LIB\!\!\!       &  \!\!\!BR+$k$NN\!\!\!       & \!\!\!FastXML\!\!\!     & \!\!\!SLEEC\!\!\!      & \!\!\!$m=256$\!\!\! & \!\!\!$m=512$\!\!\!  & \!\!\!$m=1024$\!\!\!         &   \!\!\!$m=256$\!\!\!    &    \!\!\!$m=512$\!\!\!   &  \!\!\!$m=1024$\!\!\! \\
					\hline
					\!\!\!corel5k\!\!\!             &  \!\!\!0.1150\!\!\!        & \!\!\!0.0930\!\!\!         & \!\!\!0.0530\!\!\!       & \!\!\!0.0824\!\!\!     & \!\!\!0.0475\!\!\!   & \!\!\!0.0446\!\!\!   & \!\!\!0.0659\!\!\!          & \!\!\!0.0539\!\!\!           &  \!\!\!0.0817\!\!\!   & \!\!\!0.0902\!\!\!\\
					\!\!\!nus(vlad)\!\!\!           &  \!\!\!0.1247\!\!\!        & \!\!\!0.1547\!\!\!         & \!\!\!0.1118\!\!\!       & \!\!\!0.1578\!\!\!     & \!\!\!0.1099\!\!\!   & \!\!\!0.1310\!\!\!   & \!\!\!0.1460\!\!\!          & \!\!\!0.1001\!\!\!           &  \!\!\!0.1289\!\!\!   & \!\!\!0.1443\!\!\! \\
					\!\!\!nus(bow)\!\!\!            &  \!\!\!0.0984\!\!\!        & \!\!\!0.1012\!\!\!         & \!\!\!0.0892\!\!\!       & \!\!\!0.1122\!\!\!     & \!\!\!0.0896\!\!\!   & \!\!\!0.0932\!\!\!   & \!\!\!0.0952\!\!\!          & \!\!\!0.0882\!\!\!           &  \!\!\!0.0903\!\!\!   & \!\!\!0.0920\!\!\!\\
					\!\!\!rcv1x\!\!\!               &  \!\!\!0.2950\!\!\!        & \!\!\!0.2894\!\!\!         & \!\!\!0.2367\!\!\!       & \!\!\!0.2801\!\!\!     & \!\!\!0.2063\!\!\!   & \!\!\!0.2767\!\!\!   & \!\!\!0.2813\!\!\!          & \!\!\!0.2173\!\!\!           &  \!\!\!0.2621\!\!\!   & \!\!\!0.2796\!\!\! \\
					\hline
				\end{tabular}
			\end{sc}
		\end{footnotesize}
	\end{center}
\end{table*}

\subsection{Walsh-Hadamard Sketch Guarantee}

We generalize the concept of Gaussian width to two additional measures, $S$-Gaussian width and Rademacher width:
\begin{definition}[$S$-Gaussian Width] \label{ssdef4}
	Given a closed set $ \mathcal {Y} \subseteq \mathbb{R}^{n}$ and a stochastic sketch matrix $S \in \mathbb{R}^{m \times n}$, the $S$-Gaussian width of $\mathcal {Y}$, denoted by $\omega_{S}(\mathcal {Y})$, is defined as:
	\begin{equation*}\label{def3gw}
		\begin{split}
			\omega_{S}(\mathcal {Y}) = \mathbb{E}_{g,S}[\sup_{z \in \mathcal {Y}}|\langle g,\frac{Sz}{\sqrt{m}} \rangle| ]
		\end{split}
	\end{equation*}
	where $g \thicksim N(0,\mathbf{I}_{m\times m}) $.
\end{definition}

\begin{definition}[Rademacher Width] \label{ssdef5}
	Given a closed set $ \mathcal {Y} \subseteq \mathbb{R}^{n}$, the Rademacher width  of $\mathcal {Y}$, denoted by $\Upsilon(\mathcal {Y})$, is defined as:
	\begin{equation*}\label{def3gw}
		\begin{split}
			\Upsilon(\mathcal {Y}) = \mathbb{E}_{\varpi}[\sup_{z \in \mathcal {Y}}|\langle \varpi,z\rangle| ]
		\end{split}
	\end{equation*}
	where $\varpi \in \{ \pm 1\}^n $ is an i.i.d. vector of Rademacher variables.
\end{definition}
Next, we still define $\mathcal {Y}= X\mathcal {K} \cap \mathbb{S}^{n-1}$ and state the following theorem for guaranteeing the Walsh-Hadamard sketch:

\begin{theorem}\label{subgaussiansketchtheorem3}
	Let $S \in \mathbb{R}^{m \times n}$ be a stochastic Walsh-Hadamard sketch matrix, $c_1$, $c_2$ and $c_3$ be universal constants. Given any $\delta \in(0,1)$ and $ m = \mathcal{O}( (\frac{c_1}{\delta})^2 (\Upsilon(\mathcal {Y}) +\sqrt{6log(n)})^2 \omega^2_{S}(\mathcal {Y}) )$,
	then with probability at least $1- 6q\big(\frac{c_2}{(mn)^2} + c_2 e^{-\frac{c_3 m \delta^2 }{\Upsilon(\mathcal {Y})^2 +log(nm)}}\big)$, $\hat{V}$ is a $\delta$-optimality approximation solution.
\end{theorem}

\textbf{Remark.} An additional term $(\Upsilon(\mathcal {Y}) +\sqrt{6log(n)})^2$ appears in the sketch size, so the required sketch size for the Walsh-Hadamard sketch is larger than that required for the $\sigma$-Subgaussian sketch. However, the potentially larger sketch size is offset by the much lower cost of matrix multiplications via the stochastic Walsh-Hadamard sketch matrix. Theorem \ref{subgaussiansketchtheorem3} guarantees that the stochastic Walsh-Hadamard sketch method is also able to construct a small sketch of the full data set for the fast optimization of problem \ref{2}, while preserving the $\delta$-optimality of the solution.

\subsection{Generalization Error Bound}
\label{GEB}

This subsection provides the generalization error bound of the multi-label classification problem using our proposed two stochastic sketch models. Because our results can be applied to two models, we simply call our stochastic sketch models SS.
Assume our model is characterized by a distribution $\mathcal {D}$ on the space of inputs and labels $ \mathcal {X} \times \{0,1\}^q$, where $\mathcal {X} \subseteq \mathbb{R}^p$. Let a sample $\{(x^{(j)},y^{(j)})\}$ be drawn i.i.d. from the distribution $\mathcal {D}$, where $y^{(j)} \in \{0,1\}^q$ $(j \in \{1,\ldots, n\})$ are the ground truth label vectors. Assume $n$ samples $D=\{(x^{(1)},y^{(1)}),\cdots, (x^{(n)},y^{(n)})\}$ are drawn i.i.d. $n$ times from the distribution $\mathcal {D}$, which is denoted by $D \sim \mathcal {D}^n$. For two inputs $x^{(z)},x^{(j)}$ in $\mathcal {X}$, we define $d(x^{(z)},x^{(j)})= ||x^{(z)}  -  x^{(j)}||_2$ as the Euclidean metric in the original input space and $d_{pro}(x^{(z)},x^{(j)})= \|\hat{V}'x^{(z)} -  \hat{V}'x^{(j)}\|_2$ as the metric in the embedding input space. Let $h_{knn_i}^D(x)$ represent the prediction of the $i$-th label for input $x$ using our model SS-$k$NN, which is trained on $D$. The performance of SS-$k$NN: $(h_{knn_1}^D(\cdot),\cdots,h_{knn_q}^D(\cdot)): \mathcal {X} \rightarrow \{0,1\}^q $ is then measured in terms of its generalization error, which is its expected loss on a new example $(x,y)$ drawn according to $\mathcal {D}$:
\begin{equation}\label{generalizederror}
	\begin{split}
		E_{D \sim \mathcal {D}^n,(x,y)\sim \mathcal {D}} \Big( \sum\limits_{i=1}^q \ell (y_i,h_{knn_i}^D(x))\Big)
	\end{split}
\end{equation}
where $y_i$ means the $i$-th label and $\ell (y_i,h_{knn_i}^D(x))$ represents the loss function for the $i$-th label. We define the loss function as follows for the analysis.
\begin{equation}\label{lossfunction}
	\begin{split}
		\ell (y_i,h_{knn_i}^D(x)) = P(y_i \neq h_{knn_i}^D(x))
	\end{split}
\end{equation}
For the $i$-th label, we define the function as follows:
\begin{equation}\label{eta}
	\begin{split}
		\nu_j^i(x)=P(y_i=j |x),j\in\{0,1\}.
	\end{split}
\end{equation}
The Bayes optimal classifier $b^*$ for the $i$-th label is defined as
\begin{equation}\label{bayesoptimal}
	\begin{split}
		b_i^*(x)= \arg\max _{j \in \{0,1\}}  \nu_j^i(x)
	\end{split}
\end{equation}
Before deriving our results, we first present several important definitions and theorems.
\begin{definition}[Covering Numbers, \cite{journals/tit/Shawe-TaylorBWA98}] \label{def1}
	Let $(\mathcal {X},d)$ be a metric space, $A$ be a subset of $\mathcal {X}$ and $\varepsilon >0$. A set $B \subseteq X$ is an $\varepsilon$-cover for $A$, if for every $a\in A$, there exists $b \in B$ such that $d(a,b)< \varepsilon$. The $\varepsilon$-covering number of $A$, $\mathcal {N}(\varepsilon, A,d)$, is the minimal cardinality of an $\varepsilon$-cover for $A$ (if there is no such finite cover then it is defined as $\infty$).
\end{definition}
\begin{definition}[Doubling Dimension, \cite{DBLP:conf/soda/KrauthgamerL04}]\label{def2}
	Let $(\mathcal {X},d)$ be a metric space, and let $\bar{\lambda}$ be the smallest value such that every ball in $\mathcal {X}$ can be covered by $\bar{\lambda}$ balls of half the radius. The doubling dimension of $\mathcal {X}$ is defined as : $ddim(\mathcal {X})= \log_2(\bar{\lambda})$.
\end{definition}
\begin{theorem}[\cite{DBLP:conf/soda/KrauthgamerL04}] \label{the2}
	Let $(\mathcal {X},d)$ be a metric space. The diameter of $\mathcal {X}$ is defined as $diam( \mathcal {X})= \sup\limits_{x,x' \in \mathcal {X}} d(x,x')$. The $\varepsilon$-covering number of $\mathcal {X}$, $\mathcal {N}(\varepsilon, \mathcal {X},d)$, is bounded by:
	\begin{equation}\label{coverbound}
		\begin{split}
			\mathcal {N}(\varepsilon, \mathcal {X},d) \leq \Big( \frac{2diam( \mathcal {X})}{\varepsilon} \Big)^{ddim(\mathcal {X})}
		\end{split}
	\end{equation}
\end{theorem}

\begin{table*}
	\caption{The training time (in second) on the various data sets.}
	\label{time}
	\begin{center}
		\begin{footnotesize}
			\begin{sc}
				\begin{tabular}{lcccccccccr}
					\hline
					&                &                 &             &            &\multicolumn{3}{c}{SS+GAU}        &          \multicolumn{3}{c}{SS+WH}    \\
					\!\!\!Data Set\!\!\!            &   \!\!\!BR+LIB\!\!\!       &  \!\!\!BR+$k$NN\!\!\!       & \!\!\!FastXML\!\!\!     & \!\!\!SLEEC\!\!\!      & \!\!\!$m=256$\!\!\! & \!\!\!$m=512$\!\!\!  & \!\!\!$m=1024$\!\!\!    &   \!\!\!$m=256$\!\!\!    &    \!\!\!$m=512$\!\!\!   &  \!\!\!$m=1024$\!\!\! \\
					\hline
					\!\!\!corel5k\!\!\!             &  \!\!\!7.198\!\!\!         &  \!\!\!0.678\!\!\!         & \!\!\!4.941\!\!\!        & \!\!\!736.670\!\!\!     & \!\!\!0.196\!\!\!    & \!\!\!0.218\!\!\!        & \!\!\!0.366\!\!\!           &\!\!\!0.119\!\!\!             &  \!\!\!0.197\!\!\!        &\!\!\!0.239\!\!\!   \\
					\!\!\!nus(vlad)\!\!\!           &  \!\!\!222.21\!\!\!        &  \!\!\!179.04\!\!\!        & \!\!\!715.86\!\!\!       & \!\!\!9723.49\!\!\!    & \!\!\!25.29\!\!\!    & \!\!\!51.68\!\!\!    & \!\!\!93.97\!\!\!           & \!\!\!11.87\!\!\!            &  \!\!\!20.22\!\!\!    & \!\!\!33.04\!\!\! \\
					\!\!\!nus(bow)\!\!\!            &  \!\!\!511.83\!\!\!        & \!\!\!351.64\!\!\!         & \!\!\!1162.53\!\!\!      & \!\!\!11391.54\!\!\!   & \!\!\!52.05\!\!\!    & \!\!\!72.65\!\!\!    & \!\!\!120.37\!\!\!          & \!\!\!25.41\!\!\!            &  \!\!\!34.32\!\!\!    & \!\!\!48.85\!\!\!\\
					\!\!\!rcv1x\!\!\!              &  \!\!\!22607.53\!\!\!      & \!\!\!353.42\!\!\!         & \!\!\!1116.05\!\!\!      & \!\!\!78441.93\!\!\!   & \!\!\!72.53\!\!\!    & \!\!\!114.55\!\!\!   & \!\!\!144.17\!\!\!          & \!\!\!48.88\!\!\!            &  \!\!\!55.94\!\!\!    & \!\!\!72.22\!\!\!\\
					\hline
				\end{tabular}
			\end{sc}
		\end{footnotesize}
	\end{center}
\end{table*}

We provide the following generalization error bound for SS-1NN:
\begin{theorem}\label{the3}
	Given a metric space $(\mathcal {X}, d_{pro})$, assume function $\nu^i : \mathcal {X} \rightarrow [0,1]$ is Lipschitz with constant $L$ with respect to the sup-norm for each label. Suppose $\mathcal {X}$ has a finite doubling dimension: $ddim(\mathcal {X}) = \mathbb{D} < \infty$ and $diam( \mathcal {X}) =1$. Let $D=\{(x^{(1)},y^{(1)}),\cdots, (x^{(n)},y^{(n)})\}$ and $(x,y)$ be drawn i.i.d. from the distribution $\mathcal {D}$. Then, we have
	\begin{equation}\label{lm1nnbound}
		\begin{split}
			&E_{D \sim \mathcal {D}^n,(x,y)\sim \mathcal {D}} \Big( \sum\limits_{i=1}^q   P(y_i \neq h_{1nn_i}^D(x))\Big)\\
			&\leq \sum\limits_{i=1}^q 2P(b_i^*(x)\neq y_i)  + \frac{3qL||\hat{V}||_F}{n^{1/(\mathbb{D}+1)}}
		\end{split}
	\end{equation}
\end{theorem}

Inspired by Theorem 19.5 in \cite{Shalev-Shwartz:2014:UML:2621980}, we derive the following lemma for SS-$k$NN:
\begin{lemma}\label{lemma1}
	Given metric space $(\mathcal {X}, d_{pro})$, assume function $\nu^i : \mathcal {X} \rightarrow \{0,1\}$ is Lipschitz with constant $L$ with respect to the sup-norm for each label. Suppose $\mathcal {X}$ has a finite doubling dimension: $ddim(\mathcal {X}) = \mathbb{D} < \infty$ and $diam( \mathcal {X}) =1$. Let $D=\{(x^{(1)},y^{(1)}),\cdots, (x^{(n)},y^{(n)})\}$ and $(x,y)$ be drawn i.i.d. from the distribution $\mathcal {D}$. Then, we have
	\begin{equation}\label{lmknnbound}
		\begin{split}
			& E_{D \sim \mathcal {D}^n,(x,y)\sim \mathcal {D}} \Big( \sum\limits_{i=1}^q   P(y_i \neq h_{knn_i}^D(x))\Big)\\
			&\leq \sum\limits_{i=1}^q (1+\sqrt{8/k})P(b_i^*(x)\neq y_i) \!\!+\!\! \frac{q(6L||\hat{V}||_F+k)}{n^{1/(\mathbb{D}+1)}} \\
		\end{split}
	\end{equation}
\end{lemma}

The following corollary reveals important statistical properties of SS-1NN and SS-$k$NN.
\begin{corollary}\label{corollary1}
	As $n$ goes to infinity, the error of the SS-1NN and SS-$k$NN converges to the sum of twice the Bayes error and $1+\sqrt{8/k}$ times Bayes error over the labels, respectively.
\end{corollary}

\section{Experiment}
\label{Exsec}
\begin{figure*}
	\begin{center}
		\includegraphics[scale=0.6]{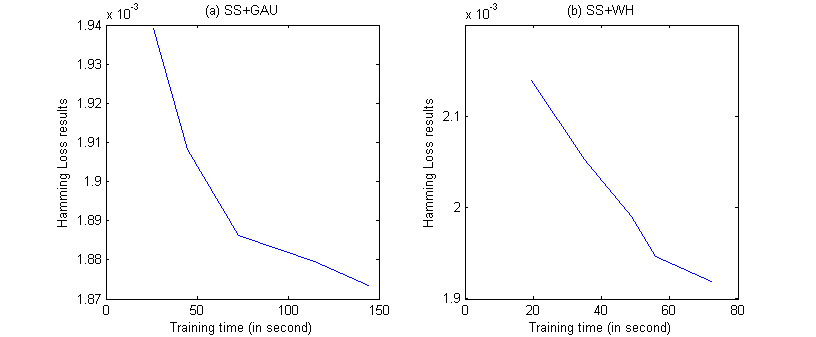}
	\end{center}
	\vspace{-14pt}
	\caption{\label{Experimentresults} Experiment results of SS+GAU and SS+WH on rcv1x data set.}
\end{figure*}

\makeatletter
\def\hlinew#1{%
	\noalign{\ifnum0=`}\fi\hrule \@height #1 \futurelet
	\reserved@a\@xhline}
\subsection{Data Sets and Baselines}

We abbreviate our proposed stochastic $\sigma$-Subgaussian sketch and stochastic Walsh-Hadamard sketch to SS+GAU and SS+WH, respectively. In the experiment, we set the entries in the $\sigma$-Subgaussian sketch matrix as  i.i.d standard Gaussian entries.
This section evaluates the performance of the proposed methods on four data sets: corel5k, nus(vlad), nus(bow) and rcv1x. The statistics of these data sets are presented in website\footnote{http://mulan.sourceforge.net}.
We compare SS+GAU and SS+WH  with several state-of-the-art methods, as follows.
\begin{itemize}
	\item BR \cite{Tsoumakas2010}: We implement two base classifiers for BR. The first uses linear classification/regression package LIBLINEAR \cite{liblinear} with $l_2$-regularized square hinge loss as the base classifier. We simply call this baseline BR+LIB. The second uses $k$NN as the base classifier. We simply call this baseline BR+$k$NN and count the $k$NN search time as the training time.
	\item FastXML \cite{Prabhu14}: An advanced tree-based multi-label classifier.
	\item SLEEC \cite{Bhatia2015}: A state-of-the-art embedding method, which is based on sparse local embeddings for large-scale multi-label classification. We use solvers of FastXML and SLEEC provided by the respective authors with default parameters.
\end{itemize}

Following the similar settings in \cite{journals/pr/ZhangZ07} and \cite{Bhatia2015}, we set $k=10$ for the $k$NN search in all $k$NN based methods.
The sketch size $m$ is chosen in a range of $\{64, 128, 256, 512, 1024\}$.
Following \cite{DBLP:conf/nips/ChenL12}, \cite{conf/icml/ZhangS12} and \cite{DBLP:conf/pkdd/GuoS13}, we consider the Hamming Loss and Example-F1 measures to evaluate the prediction performance of all the methods.  The smaller the value of the Hamming Loss, the better the performance, while the larger the value of Example-F1, the better the performance.


\subsection{Results}

Figure \ref{Experimentresults} shows that with the increasing sketch size, the training time of SS+GAU and SS+WH rise, while the prediction performance of SS+GAU and SS+WH becomes better. The results verify our theoretical analysis. The Hamming Loss, Example-F1 and training time comparisons of various methods on corel5k, nus(vlad), nus(bow) and rcv1x data sets are shown in Table~\ref{Hamres}, Table~\ref{ExampleF1} and Table~\ref{time}, respectively.
From Tables ~\ref{Hamres}, ~\ref{ExampleF1} and \ref{time}, we can see that:
\begin{itemize}
	\item BR and SLEEC usually achieve better results, which is consistent with the empirical results in \cite{Bhatia2015} and \cite{BinaryrelevanceBRmethodclassifier}. However, SLEEC is the slowest method compared to other baselines.
	\item Because we perform the optimization only on a small sketch of the full data set, our proposed methods are significantly faster than BR and state-of-the-art embedding approaches. Moreover, we can maintain competitive prediction performance by setting an appropriate sketch size. The empirical results illustrate our theoretical studies.
\end{itemize}

\section{Conclusion}


This paper carefully constructs stochastic $\sigma$-Subgaussian sketch and Walsh-Hadamard sketch for multi-label classification. From an algorithmic perspective, we show that we can obtain answers
that are approximately as good as the exact answer for BR. From a statistical learning perspective, we also provide the generalization error bound of multi-label classification using our proposed stochastic sketch model. 
Lastly, our empirical studies corroborate our theoretical findings, and demonstrate the superiority of the proposed methods.

\section*{Supplementary: The Proof of Important Theorems and Lemmas}

\subsection{Proof of Theorem 1}
We first present the following lemma, which is derived from \cite{ReconstructionandSubgaussianOperators} and \cite{DBLP:journals/tit/PilanciW15}.
\begin{customlemma}{1}\label{proofTh1lemma1}
	Let $S \in \mathbb{R}^{m \times n}$ be a stochastic $\sigma$-Subgaussian sketch matrix. Then there are universal constants $c_1$ and $c_2$ such that for any subset $ \mathcal {Y} \subseteq \mathbb{S}^{n-1}$, any $u \in \mathbb{S}^{n-1}$ and $\delta \in(0,1)$, we have
	\begin{equation}\label{proofTh1lemma1eq1}
		\begin{split}
			\sup_{z\in \mathcal {Y}  } | z'\mathscr{S} z| \leq \frac{c_1}{\sqrt{m}} \omega(\mathcal {Y}) + \delta
		\end{split}
	\end{equation}
	with probability at least $1 - e^{-\frac{c_2 m \delta^2 }{\sigma^4}}$, and we have
	\begin{equation}\label{proofTh1lemma1eq2}
		\begin{split}
			\sup_{z\in \mathcal {Y}  } | z' \mathscr{S} u| \leq \frac{5c_1}{\sqrt{m}} \omega(\mathcal {Y}) + 3\delta
		\end{split}
	\end{equation}
	with probability at least $1 - 3e^{-\frac{c_2 m \delta^2 }{\sigma^4}}$, where $\mathscr{S}= S'S - \mathbf{I}_{n\times n}$.
\end{customlemma}

\begin{customthm}{1}\label{subgaussiansketchtheorem2}
	Let $S \in \mathbb{R}^{m \times n}$ be a stochastic $\sigma$-Subgaussian sketch matrix, $c_1$ and $c_2$ be universal constants. Given any $\delta \in(0,1)$ and $m =\mathcal{O}( (\frac{c_1}{\delta})^2 \omega^2(\mathcal {Y})) $,
	then with probability at least $1- 6qe^{-\frac{c_2 m \delta^2 }{\sigma^4}}$, $\hat{V}$ is a $\delta$-optimality approximation solution.
\end{customthm}
\begin{proof}
	Let $V^*_1,\cdots,V^*_q$, $\hat{V}_1,\cdots,\hat{V}_q$  and $Y_1,\cdots,Y_q$ be $q$ columns of matrix $V^*$, $\hat{V}$ and $Y$, respectively. Then $||XV^* - Y||^2_F$ and $||SX\hat{V} - SY||^2_F$ can be decomposed to $||XV^* - Y||^2_F= \sum^q_{i=1}||XV^*_i - Y_i||^2_2$ and $||SX\hat{V} - SY||^2_F=\sum^q_{i=1} ||SX\hat{V}_i - SY_i||^2_2$.  Next, we study the relationship between $||XV^*_i - Y_i||^2_2$ and $||SX\hat{V}_i - SY_i||^2_2$.
	We define $M=\hat{V}_i-V^*_i$. According to Definition 3 in the main paper, we know that $M$ belongs to the tangent cone of $\mathcal {C}$ at $V^*_i$.
	
	Because $V^*_i \in \arg\min_{r\in \mathbb{R}^p}||Xr - Y_i||^2_2 $, we have $||XV^*_i - Y_i||^2_2  \leq  ||X\hat{V}_i - Y_i||^2_2 = ||XV^*_i - Y_i||^2_2
	+ 2\langle XV^*_i - Y_i,XM \rangle +||XM||^2_2$.
	Then, we get:
	\begin{equation}\label{proofTh1eq1}
		\begin{split}
			2\langle XV^*_i - Y_i,XM \rangle +||XM||^2_2 \geq 0
		\end{split}
	\end{equation}
	As $\hat{V}_i \in \arg\min_{r\in \mathbb{R}^p}||SXr - SY_i||^2_2 $, we have $||SXV^*_i - SY_i||^2_2 \geq  ||SX\hat{V}_i - SY_i||^2_2 = ||SXV^*_i - SY_i||^2_2
	+2\langle SXV^*_i - SY_i,SXM \rangle +||SXM||^2_2$.
	Then, we get $||SXM||^2_2 \leq - 2\langle SXV^*_i - SY_i,SXM \rangle \leq 2 ||SXV^*_i - SY_i||_2 ||SXM||_2$ and
	\begin{equation}\label{proofTh1eq2}
		\begin{split}
			||SXM||_2 \leq 2 ||SXV^*_i - SY_i||_2
		\end{split}
	\end{equation}
	We derive the following:
	\begin{equation*}
		\begin{split}
			&||SX\hat{V}_i - SY_i||^2_2\\
			=&||SXV^*_i - SY_i||^2_2  \!\!+\!\!||SXM||^2_2\!\!+\!\! 2\langle SXV^*_i - SY_i,SXM \rangle\\
			=&||SXV^*_i - SY_i||^2_2  +||XM||^2_2 + \langle XM ,\mathscr{S} XM \rangle\\
			&+ 2\langle XV^*_i - Y_i,\mathscr{S} XM \rangle + 2\langle XV^*_i - Y_i,XM \rangle
		\end{split}
	\end{equation*}
	By using Lemma \ref{proofTh1lemma1}, with probability at least $1 - 4e^{-\frac{c_2 m \delta^2 }{\sigma^4}}$, we have
	\begin{equation*}
		\begin{split}
			&||SX\hat{V}_i - SY_i||^2_2\\
			\leq &||SXV^*_i \!\!-\!\! SY_i||^2_2  +||XM||^2_2 (1+ \frac{c_1}{\sqrt{m}} \omega(\mathcal {Y}) + \delta) \\
			&+ 2||XV^*_i - Y_i||_2 ||XM||_2 (1+\frac{5c_1}{\sqrt{m}} \omega(\mathcal {Y}) + 3\delta)\\
		\end{split}
	\end{equation*}
	where $\mathcal {Y}= X\mathcal {K} \cap \mathbb{S}^{n-1}$. Given $\gamma >0$, we have $2||XV^*_i - Y_i||_2 ||XM||_2 \leq \gamma ||XV^*_i - Y_i||^2_2 + 1/\gamma ||XM||^2_2$. For the sake of clarity, we define $\psi=1+ \frac{5c_1}{\sqrt{m}} \omega(\mathcal {Y}) + 3\delta$ and $\varphi=1+ \frac{c_1}{\sqrt{m}} \omega(\mathcal {Y}) + \delta$, and then substitute them to the above expression, with probability at least $1 - 4e^{-\frac{c_2 m \delta^2 }{\sigma^4}}$, we have
	\begin{equation}\label{proofTh1eq3}
		\begin{split}
			&||SX\hat{V}_i - SY_i||^2_2\\
			\leq &||SXV^*_i \!\!-\!\! SY_i||^2_2 \!\! +  \gamma \psi ||XV^*_i - Y_i||^2_2 \!\!+ \!\!(\frac{\psi}{\gamma}+\varphi)||XM||^2_2\\
		\end{split}
	\end{equation}
	Clearly, we have $\omega( \frac{XV^*_i -Y_i}{||XV^*_i -Y_i||_2}) \leq \omega(\mathcal {Y}) $.  By using Lemma \ref{proofTh1lemma1}, with probability at least $1 - e^{-\frac{c_2 m \delta^2 }{\sigma^4}}$, we have
	\begin{equation}\label{proofTh1eq4}
		\begin{split}
			&||SXV^*_i - SY_i||^2_2 \\
			= & ||XV^*_i - Y_i||^2_2 + \langle XV^*_i - Y_i,\mathscr{S}(XV^*_i - Y_i) \rangle\\
			\leq & ||XV^*_i - Y_i||^2_2 (1+ \frac{c_1}{\sqrt{m}} \omega(\frac{XV^*_i -Y_i}{||XV^*_i -Y_i||_2}) + \delta)\\
			\leq & ||XV^*_i - Y_i||^2_2 \varphi
		\end{split}
	\end{equation}
	By using Lemma \ref{proofTh1lemma1}, with probability at least $1 - e^{-\frac{c_2 m \delta^2 }{\sigma^4}}$, we have $||SXM||^2_2=||XM||^2_2 + \langle XM,\mathscr{S}XM\rangle \geq ||XM||^2_2 (1- \frac{c_1}{\sqrt{m}} \omega(\mathcal {Y}) - \delta)= ||XM||^2_2 (2 - \varphi)$. By using Eq.(\ref{proofTh1eq2}), with probability at least $1 - e^{-\frac{c_2 m \delta^2 }{\sigma^4}}$, we have
	\begin{equation}\label{proofTh1eq5}
		\begin{split}
			||XM||^2_2
			\leq  \frac{||SXM||^2_2}{2 - \varphi}
			\leq  4\frac{||SXV^*_i - SY_i||^2_2 }{2 - \varphi}
		\end{split}
	\end{equation}
	Eq.(\ref{proofTh1eq3}), Eq.(\ref{proofTh1eq4}) and Eq.(\ref{proofTh1eq5}) imply that, with probability at least $1 - 6e^{-\frac{c_2 m \delta^2 }{\sigma^4}}$, we have
	\begin{equation}\label{proofTh1eq6}
		\begin{split}
			&||SX\hat{V}_i - SY_i||^2_2\\
			\leq & (1 + 4\frac{\frac{\psi}{\gamma}\!\!+\!\!\varphi }{2 - \varphi})||SXV^*_i \!\!-\!\! SY_i||^2_2 \!\! +    \gamma \psi ||XV^*_i  \!\!-  \!\!Y_i||^2_2\\
			\leq & (1 \!\!+ \!4\frac{\frac{\psi}{\gamma}\!\!+\!\!\varphi }{2 \!\!-\!\! \varphi})\varphi ||XV^*_i\!\! -\!\! Y_i||^2_2\!\!+  \gamma \psi ||XV^*_i \!\!-\!\! Y_i||^2_2\\
			\leq & (\varphi - 4\frac{\psi}{\gamma}- 4\varphi +  \gamma \psi ) ||XV^*_i - Y_i||^2_2\\
		\end{split}
	\end{equation}
	By setting $\gamma=4$, with probability at least $1 - 6e^{-\frac{c_2 m \delta^2 }{\sigma^4}}$, we have
	\begin{equation}\label{proofTh1eq7}
		\begin{split}
			&||SX\hat{V}_i - SY_i||^2_2\\
			\leq & (3\psi - 3\varphi ) ||XV^*_i - Y_i||^2_2\\
			= & (\frac{12 c_1}{\sqrt{m}} \omega(\mathcal {Y}) + 6 \delta ) ||XV^*_i - Y_i||^2_2\\
		\end{split}
	\end{equation}
	Eq.(\ref{proofTh1eq7}) implies that, with probability at least $1 - 6qe^{-\frac{c_2 m \delta^2 }{\sigma^4}}$, we have
	\begin{equation}\label{proofTh1eq8}
		\begin{split}
			||SX\hat{V} \!\!-\!\! SY||^2_F
			\leq (\frac{12 c_1}{\sqrt{m}} \omega(\mathcal {Y}) + 6 \delta) ||XV^* \!\!-\!\! Y||^2_F
		\end{split}
	\end{equation}

	By using Eq.(\ref{proofTh1eq1}) and Lemma \ref{proofTh1lemma1} again, with probability at least $1 - 4e^{-\frac{c_2 m \delta^2 }{\sigma^4}}$, we have
	\begin{equation*}
		\begin{split}
			&||SX\hat{V}_i - SY_i||^2_2\\
			\geq &||SXV^*_i \!\!-\!\! SY_i||^2_2 \!\! + \!\!\langle XM ,\mathscr{S} XM \rangle \!\!+ \!\! 2\langle XV^*_i\!\! - \!\!Y_i,\mathscr{S} XM \rangle \\
			\geq &||SXV^*_i \!\!-\!\! SY_i||^2_2  -  ||XM||^2_2(\frac{c_1}{\sqrt{m}} \omega(\mathcal {Y}) + \delta) \\
			&- 2||XV^*_i - Y_i||_2 ||XM||_2 (\frac{5c_1}{\sqrt{m}} \omega(\mathcal {Y}) + 3\delta)
		\end{split}
	\end{equation*}
	We define $\hat{\psi}= \frac{5c_1}{\sqrt{m}} \omega(\mathcal {Y}) + 3\delta$ and $\hat{\varphi}= \frac{c_1}{\sqrt{m}} \omega(\mathcal {Y}) + \delta$, and then substitute them to the above expression, with probability at least $1 - 4e^{-\frac{c_2 m \delta^2 }{\sigma^4}}$, we have
	\begin{equation}\label{proofTh1neweq3}
		\begin{split}
			&||SX\hat{V}_i - SY_i||^2_2\\
			\geq &||SXV^*_i \!\!-\!\! SY_i||^2_2 \!\! -  \gamma \hat{\psi} ||XV^*_i - Y_i||^2_2 \!\!- \!\!(\frac{\hat{\psi}}{\gamma}+\hat{\varphi})||XM||^2_2\\
		\end{split}
	\end{equation}
	By using Lemma \ref{proofTh1lemma1} again, with probability at least $1 - e^{-\frac{c_2 m \delta^2 }{\sigma^4}}$, we have
	\begin{equation}\label{proofTh1neweq4}
		\begin{split}
			||SXV^*_i - SY_i||^2_2 \geq  ||XV^*_i - Y_i||^2_2 (1 -\hat{\varphi})
		\end{split}
	\end{equation}
	Similar to Eq.(\ref{proofTh1eq5}), by using Eq.(\ref{proofTh1eq2}) and Lemma \ref{proofTh1lemma1}, with probability at least $1 - e^{-\frac{c_2 m \delta^2 }{\sigma^4}}$, we have
	\begin{equation}\label{proofTh1neweq5}
		\begin{split}
			||XM||^2_2
			\leq  \frac{||SXM||^2_2}{1 - \hat{\varphi}}
			\leq  4\frac{||SXV^*_i - SY_i||^2_2 }{1 - \hat{\varphi}}
		\end{split}
	\end{equation}
	Eq.(\ref{proofTh1neweq3}), Eq.(\ref{proofTh1neweq4}) and Eq.(\ref{proofTh1neweq5}) imply that, with probability at least $1 - 6e^{-\frac{c_2 m \delta^2 }{\sigma^4}}$, we have
	\begin{equation}\label{proofTh1eq6}
		\begin{split}
			&||SX\hat{V}_i - SY_i||^2_2\\
			\geq & (1 \!\!- \!4\frac{\frac{\hat{\psi}}{\gamma}\!\!+\!\!\hat{\varphi} }{1 \!\!-\!\! \hat{\varphi}})(1 \!\!-\!\! \hat{\varphi}) ||XV^*_i\!\! -\!\! Y_i||^2_2\!\!-  \gamma \hat{\psi} ||XV^*_i \!\!-\!\! Y_i||^2_2\\
			\geq & (1 - \hat{\varphi} - 4\frac{\hat{\psi}}{\gamma}- 4\hat{\varphi} -  \gamma \hat{\psi} ) ||XV^*_i - Y_i||^2_2\\
		\end{split}
	\end{equation}
	By setting $\gamma=2$, with probability at least $1 - 6e^{-\frac{c_2 m \delta^2 }{\sigma^4}}$, we have
	\begin{equation}\label{proofTh1neweq7}
		\begin{split}
			&||SX\hat{V}_i - SY_i||^2_2\\
			\geq & (1 - 4\hat{\psi} - 5\hat{\varphi} ) ||XV^*_i - Y_i||^2_2\\
			= & (1 - \frac{25 c_1}{\sqrt{m}} \omega(\mathcal {Y}) - 17 \delta ) ||XV^*_i - Y_i||^2_2\\
		\end{split}
	\end{equation}
	Eq.(\ref{proofTh1neweq7}) implies that, with probability at least $1 - 6qe^{-\frac{c_2 m \delta^2 }{\sigma^4}}$, we have
	\begin{equation}\label{proofTh1neweq8}
		\begin{split}
			||SX\hat{V} \!\!-\!\! SY||^2_F
			\geq (1\!\! - \!\!\frac{25 c_1}{\sqrt{m}} \omega(\mathcal {Y})\!\! -\!\! 17 \delta ) ||XV^* \!\!-\!\! Y||^2_F
		\end{split}
	\end{equation}
	By rescaling $\delta$ and redefining the universal constants appropriately for Eq.(\ref{proofTh1eq8}) and Eq.(\ref{proofTh1neweq8}), we prove Theorem \ref{subgaussiansketchtheorem2}.
\end{proof}

\subsection{Proof of Theorem 2}
Let $\mathbb{O}^{n}=\{z\in \mathbb{R}^{n}| ||z||_2 \leq 1\}$ be the Euclidean ball of radius one. We present the following Lemma, which is derived from \cite{DBLP:journals/tit/PilanciW15}.
\begin{customlemma}{2}\label{proofTh2lemma1}
	Let $S \in \mathbb{R}^{m \times n}$ be a stochastic Walsh-Hadamard sketch matrix. Then there are universal constants $c_1$ and $c_2$ such that for any subset $ \mathcal {Y} \subseteq \mathbb{O}^{n}$, any $u \in \mathbb{S}^{n-1}$ and $\delta \in(0,1)$, we have
	\begin{equation}\label{proofTh2lemma1eq1}
		\begin{split}
			\sup_{z\in \mathcal {Y}  } | z'\mathscr{S} z| \leq \Phi (\mathcal {Y}) + \frac{\delta}{2}
		\end{split}
	\end{equation}
	with probability at least $1- \big(\frac{c_1}{(mn)^2} + c_1 e^{-\frac{c_2 m \delta^2 }{\Upsilon(\mathcal {Y})^2 +log(nm)}}\big)$, and we have
	\begin{equation}\label{proofTh2lemma1eq2}
		\begin{split}
			\sup_{z\in \mathcal {Y}  } | z' \mathscr{S} u| \leq 39 \Phi (\mathcal {Y}) + 3\delta
		\end{split}
	\end{equation}
	with probability at least $1- \big(3\frac{c_1}{(mn)^2} + 3 c_1 e^{-\frac{c_2 m \delta^2 }{\Upsilon(\mathcal {Y})^2 +log(nm)}}\big)$, where $\mathscr{S}= S'S - \mathbf{I}_{n\times n}$. $ \Phi (\mathcal {Y}) =8(\Upsilon(\mathcal {Y}) +\sqrt{6log(nm)}) \omega_{S}(\mathcal {Y})/\sqrt{m}$.
\end{customlemma}

\begin{customthm}{2}\label{subgaussiansketchtheorem3}
	Let $S \in \mathbb{R}^{m \times n}$ be a stochastic Walsh-Hadamard sketch matrix, $c_1$, $c_2$ and $c_3$ be universal constants. Given any $\delta \in(0,1)$ and $ m = \mathcal{O}( (\frac{c_1}{\delta})^2 (\Upsilon(\mathcal {Y}) +\sqrt{6log(n)})^2 \omega^2_{S}(\mathcal {Y}) )$,
	then with probability at least $1- 6q\big(\frac{c_2}{(mn)^2} + c_2 e^{-\frac{c_3 m \delta^2 }{\Upsilon(\mathcal {Y})^2 +log(nm)}}\big)$, $\hat{V}$ is a $\delta$-optimality approximation solution.
\end{customthm}
\begin{proof}
	The proof idea is similar to the proof of Theorem \ref{subgaussiansketchtheorem2}. We also define $V^*_1,\cdots,V^*_q$, $\hat{V}_1,\cdots,\hat{V}_q$  and $Y_1,\cdots,Y_q$ as $q$ columns of matrix $V^*$, $\hat{V}$ and $Y$, respectively. Then, we study the relationship between $||XV^*_i - Y_i||^2_2$ and $||SX\hat{V}_i - SY_i||^2_2$. Let $M=\hat{V}_i-V^*_i$.
	By using Lemma \ref{proofTh2lemma1}, with probability at least $1 - 4\big(\frac{c_1}{(mn)^2} + c_1 e^{-\frac{c_2 m \delta^2 }{\Upsilon(\mathcal {Y})^2 +log(nm)}}\big)$, we have
	\begin{equation*}
		\begin{split}
			&||SX\hat{V}_i - SY_i||^2_2\\
			\leq &||SXV^*_i \!\!-\!\! SY_i||^2_2  +  ||XM||^2_2(1+ \Phi (\mathcal {Y}) + \frac{\delta}{2}) \\
			&+ 2||XV^*_i - Y_i||_2 ||XM||_2 (1+39 \Phi (\mathcal {Y}) + 3\delta)
		\end{split}
	\end{equation*}
	where $\mathcal {Y}= X\mathcal {K} \cap \mathbb{S}^{n-1}$.
	Given $\gamma >0$, we have $2||XV^*_i - Y_i||_2 ||XM||_2 \leq \gamma ||XV^*_i - Y_i||^2_2 + 1/\gamma ||XM||^2_2$. For the sake of clarity, we redefine $\psi= 1+ 39 \Phi (\mathcal {Y}) + 3\delta$ and $\varphi= 1+ \Phi (\mathcal {Y}) + \frac{\delta}{2}$, and then substitute them to the above expression, with probability at least $1 - 4\big(\frac{c_1}{(mn)^2} + c_1 e^{-\frac{c_2 m \delta^2 }{\Upsilon(\mathcal {Y})^2 +log(nm)}}\big)$, we have
	\begin{equation}\label{proofTh2eq3}
		\begin{split}
			&||SX\hat{V}_i - SY_i||^2_2\\
			\leq &||SXV^*_i \!\!-\!\! SY_i||^2_2 \!\! +  \gamma \psi ||XV^*_i\!\! -\!\! Y_i||^2_2 \!\!+ \!\!(\frac{\psi}{\gamma}+\!\!\varphi)||XM||^2_2\\
		\end{split}
	\end{equation}
	\begin{proposition}\label{proofTh2proposition2} $\Phi( \frac{XV^*_i -Y_i}{||XV^*_i -Y_i||_2}) \leq 2 \Phi(\mathcal {Y}) $
	\end{proposition}
	\begin{proof}
		$\Upsilon(X\mathcal {K} \cap \mathbb{S}^{n-1})= \mathbb{E}_{\varpi}[\sup_{z \in X\mathcal {K} \cap \mathbb{S}^{n-1} }|\langle \varpi,z\rangle| ] \geq \mathbb{E}_{\varpi}[\sum^n_{i=1} |z_i||\varpi_i|]= \sum^n_{i=1} |z_i|\mathbb{E}_{\varpi_i}[|\varpi_i|]=\sum^n_{i=1} |z_i| \geq \sum^n_{i=1} |z_i|^2 =1 $. $\Upsilon(\{\frac{XV^*_i -Y_i}{||XV^*_i -Y_i||_2}\})= \mathbb{E}_{\varpi}[\sup_{z \in \{\frac{XV^*_i -Y_i}{||XV^*_i -Y_i||_2}\}}|\langle \varpi,z\rangle| ] \leq \mathbb{E}_{\varpi}[||\varpi||_2||\frac{XV^*_i -Y_i}{||XV^*_i -Y_i||_2}||_2] =1$. Then, we get $\Upsilon(\{\frac{XV^*_i -Y_i}{||XV^*_i -Y_i||_2}\}) \leq 2 \Upsilon(X\mathcal {K} \cap \mathbb{S}^{n-1})$. Clearly, we have $\omega_{S}(\{\frac{XV^*_i -Y_i}{||XV^*_i -Y_i||_2}\}) \leq \omega_{S}(\mathcal {Y})$. Combining these yields the result.
	\end{proof}
	By using Lemma \ref{proofTh2lemma1} and Proposition \ref{proofTh2proposition2}, with probability at least $1 - \big(\frac{c_1}{(mn)^2} + c_1 e^{-\frac{c_2 m \delta^2 }{\Upsilon(\mathcal {Y})^2 +log(nm)}}\big)$, we have
	\begin{equation}\label{proofTh2eq4}
		\begin{split}
			&||SXV^*_i - SY_i||^2_2 \\
			= & ||XV^*_i - Y_i||^2_2 + \langle XV^*_i - Y_i,\mathscr{S}(XV^*_i - Y_i) \rangle\\
			\leq & ||XV^*_i - Y_i||^2_2 (1+ \Phi (\frac{XV^*_i -Y_i}{||XV^*_i -Y_i||_2}) + \frac{\delta}{2})\\
			\leq & ||XV^*_i - Y_i||^2_2 (1 + 2 \Phi(\mathcal {Y}) + \frac{\delta}{2})
		\end{split}
	\end{equation}
	By using Lemma \ref{proofTh2lemma1} and Eq.(\ref{proofTh1eq2}), with probability at least $1 - \big(\frac{c_1}{(mn)^2} + c_1 e^{-\frac{c_2 m \delta^2 }{\Upsilon(\mathcal {Y})^2 +log(nm)}}\big)$, we have
	\begin{equation}\label{proofTh2eq5}
		\begin{split}
			||XM||^2_2
			\leq  \frac{||SXM||^2_2}{2 - \varphi}
			\leq  4\frac{||SXV^*_i - SY_i||^2_2 }{2 - \varphi}
		\end{split}
	\end{equation}
	Eq.(\ref{proofTh2eq3}), Eq.(\ref{proofTh2eq4}) and Eq.(\ref{proofTh2eq5}) imply that, with probability at least $1 - 6\big(\frac{c_1}{(mn)^2} + c_1 e^{-\frac{c_2 m \delta^2 }{\Upsilon(\mathcal {Y})^2 +log(nm)}}\big)$, we have
	\begin{equation}\label{proofTh2eq6}
		\begin{split}
			&||SX\hat{V}_i - SY_i||^2_2\\
			\leq & (1 + 4\frac{\frac{\psi}{\gamma}\!\!+\!\!\varphi }{2 - \varphi})||SXV^*_i \!\!-\!\! SY_i||^2_2 \!\!+  \gamma \psi ||XV^*_i \!\!- \!\!Y_i||^2_2\\
			\leq & (1 \!\!+ 2 \Phi(\mathcal {Y}) \!+ \frac{\delta}{2}\! - \!4\frac{\psi}{\gamma}\!-\! 4\varphi \!+ \! \gamma \psi ) ||XV^*_i \!\!-\!\! Y_i||^2_2\\
		\end{split}
	\end{equation}
	By setting $\gamma=4$, with probability at least $1 - 6\big(\frac{c_1}{(mn)^2} + c_1 e^{-\frac{c_2 m \delta^2 }{\Upsilon(\mathcal {Y})^2 +log(nm)}}\big)$, we have
	\begin{equation}\label{proofTh2eq7}
		\begin{split}
			||SX\hat{V}_i \!\!-\!\! SY_i||^2_2 \leq (1 \!\!+\!\! 115 \Phi(\mathcal {Y})\!\!+ \frac{15}{2} \delta ) ||XV^*_i \!\!-\!\! Y_i||^2_2 \\
		\end{split}
	\end{equation}
	Eq.(\ref{proofTh2eq7}) implies that, with probability at least $1 - 6q\big(\frac{c_1}{(mn)^2} + c_1 e^{-\frac{c_2 m \delta^2 }{\Upsilon(\mathcal {Y})^2 +log(nm)}}\big)$, we have
	\begin{equation}\label{proofTh2eq8}
		\begin{split}
			||SX\hat{V} \!\!-\!\! SY||^2_F
			\leq (1 \!\!+\!\! 115 \Phi(\mathcal {Y}) \!\!+ \frac{15}{2} \delta ) ||XV^* \!\!- \!\!Y||^2_F
		\end{split}
	\end{equation}

	By using Eq.(\ref{proofTh1eq1}) and Lemma \ref{proofTh2lemma1} again, with probability at least $1 - 4\big(\frac{c_1}{(mn)^2} + c_1 e^{-\frac{c_2 m \delta^2 }{\Upsilon(\mathcal {Y})^2 +log(nm)}}\big)$, we have
	\begin{equation*}
		\begin{split}
			&||SX\hat{V}_i - SY_i||^2_2\\
			\geq &||SXV^*_i \!\!-\!\! SY_i||^2_2 \!\! + \!\!\langle XM ,\mathscr{S} XM \rangle \!\!+ \!\! 2\langle XV^*_i\!\! - \!\!Y_i,\mathscr{S} XM \rangle \\
			\geq &||SXV^*_i \!\!-\!\! SY_i||^2_2  -  ||XM||^2_2(\Phi (\mathcal {Y}) + \frac{\delta}{2}) \\
			&- 2||XV^*_i - Y_i||_2 ||XM||_2 (39 \Phi (\mathcal {Y}) + 3\delta)
		\end{split}
	\end{equation*}
	We redefine $\hat{\psi}= 39 \Phi (\mathcal {Y}) + 3\delta$ and $\hat{\varphi}= \Phi (\mathcal {Y}) + \frac{\delta}{2}$, and then substitute them to the above expression, with probability at least $1 - 4\big(\frac{c_1}{(mn)^2} + c_1 e^{-\frac{c_2 m \delta^2 }{\Upsilon(\mathcal {Y})^2 +log(nm)}}\big)$, we have
	\begin{equation}\label{proofTh2neweq3}
		\begin{split}
			&||SX\hat{V}_i - SY_i||^2_2\\
			\geq &||SXV^*_i \!\!-\!\! SY_i||^2_2 \!\! -  \gamma \hat{\psi} ||XV^*_i\!\! -\!\! Y_i||^2_2 \!\!- \!\!(\frac{\hat{\psi}}{\gamma}+\!\!\hat{\varphi})||XM||^2_2\\
		\end{split}
	\end{equation}
	By using Lemma \ref{proofTh2lemma1} and Proposition \ref{proofTh2proposition2} again, with probability at least $1 - \big(\frac{c_1}{(mn)^2} + c_1 e^{-\frac{c_2 m \delta^2 }{\Upsilon(\mathcal {Y})^2 +log(nm)}}\big)$, we have
	\begin{equation}\label{proofTh2neweq4}
		\begin{split}
			||SXV^*_i - SY_i||^2_2
			\geq  ||XV^*_i\!\! - \!\!Y_i||^2_2 (1\!\! -\!\! 2 \Phi(\mathcal {Y}) \!\!- \frac{\delta}{2})
		\end{split}
	\end{equation}
	Similar to Eq.(\ref{proofTh2eq5}), by using Lemma \ref{proofTh2lemma1} and Eq.(\ref{proofTh1eq2}), with probability at least $1 - \big(\frac{c_1}{(mn)^2} + c_1 e^{-\frac{c_2 m \delta^2 }{\Upsilon(\mathcal {Y})^2 +log(nm)}}\big)$, we have
	\begin{equation}\label{proofTh2neweq5}
		\begin{split}
			||XM||^2_2
			\leq  4\frac{||SXV^*_i - SY_i||^2_2 }{1 - \hat{\varphi}}
		\end{split}
	\end{equation}
	Eq.(\ref{proofTh2neweq3}), Eq.(\ref{proofTh2neweq4}) and Eq.(\ref{proofTh2neweq5}) imply that, with probability at least $1 - 6\big(\frac{c_1}{(mn)^2} + c_1 e^{-\frac{c_2 m \delta^2 }{\Upsilon(\mathcal {Y})^2 +log(nm)}}\big)$, we have
	\begin{equation}\label{proofTh2neweq6}
		\begin{split}
			&||SX\hat{V}_i - SY_i||^2_2\\
			\geq & (1 - 4\frac{\frac{\hat{\psi}}{\gamma}\!\!+\!\!\hat{\varphi} }{1 - \hat{\varphi}})||SXV^*_i \!\!-\!\! SY_i||^2_2 \!\!-  \gamma \hat{\psi}||XV^*_i \!\!- \!\!Y_i||^2_2\\
			\geq & (1 \!\!- 2 \Phi(\mathcal {Y}) \!- \frac{\delta}{2}\! - \!4\frac{\hat{\psi}}{\gamma}\!-\! 4\hat{\varphi} \!- \! \gamma \hat{\psi} ) ||XV^*_i \!\!-\!\! Y_i||^2_2\\
		\end{split}
	\end{equation}
	By setting $\gamma=2$, with probability at least $1 - 6\big(\frac{c_1}{(mn)^2} + c_1 e^{-\frac{c_2 m \delta^2 }{\Upsilon(\mathcal {Y})^2 +log(nm)}}\big)$, we have
	\begin{equation}\label{proofTh2neweq7}
		\begin{split}
			||SX\hat{V}_i \!\!-\!\! SY_i||^2_2 \geq (1 \!\!-\!\! 162 \Phi(\mathcal {Y})\!\! - \frac{29}{2} \delta ) ||XV^*_i \!\!-\!\! Y_i||^2_2 \\
		\end{split}
	\end{equation}
	Eq.(\ref{proofTh2neweq7}) implies that, with probability at least $1 - 6q\big(\frac{c_1}{(mn)^2} + c_1 e^{-\frac{c_2 m \delta^2 }{\Upsilon(\mathcal {Y})^2 +log(nm)}}\big)$, we have
	\begin{equation}\label{proofTh2neweq8}
		\begin{split}
			||SX\hat{V} \!\!-\!\! SY||^2_F
			\geq (1 \!\!-\!\! 162 \Phi(\mathcal {Y}) \!\!- \frac{29}{2} \delta ) ||XV^* \!\!- \!\!Y||^2_F
		\end{split}
	\end{equation}
	By rescaling $\delta$ and redefining the universal constants appropriately for Eq.(\ref{proofTh2eq8}) and Eq.(\ref{proofTh2neweq8}), we prove Theorem \ref{subgaussiansketchtheorem3}.
\end{proof}

\subsection{Proof of Theorem 4}

\begin{customthm}{4}\label{the3}
	Given a metric space $(\mathcal {X}, d_{pro})$, assume function $\nu^i : \mathcal {X} \rightarrow [0,1]$ is Lipschitz with constant $L$ with respect to the sup-norm for each label. Suppose $\mathcal {X}$ has a finite doubling dimension: $ddim(\mathcal {X}) = \mathbb{D} < \infty$ and $diam( \mathcal {X}) =1$. Let $D=\{(x^{(1)},y^{(1)}),\cdots, (x^{(n)},y^{(n)})\}$ and $(x,y)$ be drawn i.i.d. from the distribution $\mathcal {D}$. Then, we have
	\begin{equation}\label{lm1nnbound}
		\begin{split}
			&E_{D \sim \mathcal {D}^n,(x,y)\sim \mathcal {D}} \Big( \sum\limits_{i=1}^q   P(y_i \neq h_{1nn_i}^D(x))\Big)\\
			&\leq \sum\limits_{i=1}^q 2P(b_i^*(x)\neq y_i)  + \frac{3qL||\hat{V}||_F}{n^{1/(\mathbb{D}+1)}}
		\end{split}
	\end{equation}
\end{customthm}

\begin{proof}
	\begin{equation}\label{proof1}
		\begin{split}
			&E_{D \sim \mathcal {D}^n,(x,y)\sim \mathcal {D}} \Big( \sum\limits_{i=1}^q P(y_i \neq h_{1nn_i}^D(x)) \Big)  \\
			= & \sum\limits_{i=1}^q  E_{D \sim \mathcal {D}^n,(x,y)\sim \mathcal {D}} \Big( P(y_i \neq h_{1nn_i}^D(x))\Big)
		\end{split}
	\end{equation}
	Now, we focus on $P (y_i \neq h_{1nn_i}^D(x))$ for the $i$-th label. Given $x,x' \in \mathcal {X}$, due to $\nu^i(\cdot)$ is Lipschitz with constant $L$ with respect to the sup-norm, we have $||\nu^i(x) - \nu^i(x')||_{\infty}=  \sup\limits_{j \in \{0,1\}} |\nu_j^i(x) - \nu_j^i(x')| \leq Ld_{pro}(x,x')$ and
	\begin{equation}\label{proof2}
		\begin{split}
			&P(y_i \neq y_i'|x,x') \\
			=& P(y_i\!\!=\!\!1|x)P(y_i'\!\!=\!\!0|x')\!\!+\!\! P(y_i\!\!=\!\!0|x)P(y_i'\!\!=\!\!1|x')\\
			=& \sum\limits_{j \in \{0,1\}} \nu_j^i(x)(1-\nu_j^i(x'))\\
			\leq & \sum\limits_{j \in \{0,1\}} \nu_j^i(x)(1-\nu_j^i(x) + Ld_{pro}(x,x'))\\
			=& \sum\limits_{j \in \{0,1\}} \nu_j^i(x)(1-\nu_j^i(x)) + Ld_{pro}(x,x')
		\end{split}
	\end{equation}
	As $d_{pro}(x,x') = ||\hat{V}'x  -  \hat{V}' x'||_2 \leq ||\hat{V}||_F d(x,x') $, we get
	\begin{equation}\label{newproof2}
		\begin{split}
			&P(y_i \neq y_i'|x,x')\\
			\leq &\sum\limits_{j \in \{0,1\}} \nu_j^i(x)(1-\nu_j^i(x)) + L||\hat{V}||_F  d(x,x')
		\end{split}
	\end{equation}
	Assume $(x',y')$ is the nearest neighbor of $(x,y)$ in $D$: $(x',y')=\arg\min _{(x^{(i)},y^{(i)}) \in D } d_{pro}(x,x^{(i)})$. Then, we have $E_{D \sim \mathcal {D}^n,(x,y)\sim \mathcal {D}} \Big( P(y_i \neq h_{1nn_i}^D(x))\Big) =E_{D \sim \mathcal {D}^n,(x,y)\sim \mathcal {D}} \Big( P(y_i \neq y_i')\Big) $. Following Eq.(\ref{proof2}), we get:
	\begin{equation}\label{proof3}
		\begin{split}
			& E_{D \sim \mathcal {D}^n,(x,y)\sim \mathcal {D}} \Big( P(y_i \neq h_{1nn_i}^D(x))\Big) \\
			\leq & E_{D \sim \mathcal {D}^n,(x,y)\sim \mathcal {D}} \Big( \sum\limits_{j \in \{0,1\}} \nu_j^i(x)(1-\nu_j^i(x)) \Big)\\
			+ & L||\hat{V}||_F E_{D \sim \mathcal {D}^n,(x,y)\sim \mathcal {D}} \Big( d(x,x') \Big)
		\end{split}
	\end{equation}
	Assume the solution of $\arg\max _{j \in \{0,1\}}  \nu_j^i(x)$ is $1$. The first term of the right side of Eq.(\ref{proof3}) does not depend on $D$. Thus
	\begin{equation}\label{proof4}
		\begin{split}
			& E_{D \sim \mathcal {D}^n,(x,y)\sim \mathcal {D}} \Big( \sum\limits_{j \in \{0,1\}} \nu_j^i(x)(1-\nu_j^i(x)) \Big) \\
			=& E_{(x,y)\sim \mathcal {D}} \Big( \nu_1^i(x)(1\!\!-\!\!\nu_1^i(x))+ \nu_0^i(x)(1\!\!-\!\!\nu_0^i(x))\Big) \\
			\leq &  E_{(x,y)\sim \mathcal {D}} (1-\nu_1^i(x)) + E_{(x,y)\sim \mathcal {D}}  (\nu_0^i(x)) \\
			=&2 E_{(x,y)\sim \mathcal {D}} (1-\nu_1^i(x)) = 2 P(b_i^*(x) \neq y_i) \\
		\end{split}
	\end{equation}
	Then, we start to bound the second term of the right side of Eq.(\ref{proof3}). Let $\{C_1,\cdots,C_N\}$ be an $\varepsilon$-cover of $\mathcal {X}$ of cardinality $N=\mathcal {N}(\varepsilon, \mathcal {X},d)$. Given a sampling $D$, for $x \in C_i $ such that $D \cap C_i \neq \emptyset $, we have $d(x,x') \leq \varepsilon $, while for $x \in C_i $ such that $D \cap C_i = \emptyset $, we have $d(x,x')  \leq diam(\mathcal {X})= 1$. The expression $ [ D \cap C_i \neq \emptyset ] $ evaluates to 1 if $D \cap C_i \neq \emptyset $ is true and to 0 otherwise.  Thus, we have
	\begin{equation}\label{proof6}
		\begin{split}
			&E_{D \sim \mathcal {D}^n,(x,y)\sim \mathcal {D}} \Big( d(x,x') \Big) \\
			\leq & E_{D \sim \mathcal {D}^n}\!\!\Big( \sum\limits_{j =1} ^N P(C_j)(\varepsilon  [ D \cap C_j \!\!\neq\!\! \emptyset ] \!\!  + \!\![ D \cap C_j \!\!=\!\! \emptyset ]) \Big)\\
			\leq & \sum\limits_{j =1} ^N P(C_j) \Big( \varepsilon  E_{D \sim \mathcal {D}^n} ([ D \cap C_j \neq \emptyset ])\\
			+ & E_{D \sim \mathcal {D}^n}([ D \cap C_j = \emptyset ]) \Big)\\
		\end{split}
	\end{equation}
	Since $P(C_j)E_{D \sim \mathcal {D}^n}([ D \cap C_j = \emptyset ])=P(C_j)(1-P(C_j))^n \leq 1/en$, where $e$ is the exponent constant. This result, Eq.(\ref{proof6}) and Theorem 3 imply that
	\begin{equation}\label{proof7}
		\begin{split}
			E_{D \sim \mathcal {D}^n,(x,y)\sim \mathcal {D}} \Big(  d(x,x') \Big)
			\leq & \Big( \varepsilon  + \frac{N}{en} \Big)\\
			\leq &  \Big( \varepsilon  + \frac{1}{en}( \frac{2}{\varepsilon})^{\mathbb{D}} \Big)
		\end{split}
	\end{equation}
	By setting $\varepsilon=2n^{-\frac{1}{\mathbb{D}+1}}$, we get
	\begin{equation}\label{proof8}
		\begin{split}
			E_{D \sim \mathcal {D}^n,(x,y)\sim \mathcal {D}} \Big( d(x,x') \Big)  \leq  \frac{3}{n^{1/(\mathbb{D}+1)}}
		\end{split}
	\end{equation}
	Eq.(\ref{proof3}), Eq.(\ref{proof4}) and Eq.(\ref{proof8}) imply that:
	\begin{equation}\label{proofnew1}
		\begin{split}
			&E_{D \sim \mathcal {D}^n,(x,y)\sim \mathcal {D}} \Big( P(y_i \neq h_{1nn_i}^D(x))\Big)\\
			\leq & 2 P(b_i^*(x) \neq y_i) + \frac{3L||\hat{V}||_F}{n^{1/(\mathbb{D}+1)}}
		\end{split}
	\end{equation}
	Eq. (\ref{proof1}), Eq. (\ref{proof3}), Eq. (\ref{proof4}) and Eq. (\ref{proof8}) imply the result.
\end{proof}

\subsection{Proof of Lemma 1}

\begin{proof}
	We first focus on $E_{D \sim \mathcal {D}^n,(x,y)\sim \mathcal {D}} \Big( P(y_i \neq h_{knn_i}^D(x))\Big)$ for the $i$-th label. For each $x \in \mathcal {X}$ and training set $D=\{(x^{(1)},y^{(1)}),\cdots, (x^{(n)},y^{(n)})\}$, let $\pi_1(x),\cdots,\pi_n(x)$ be a reordering of $\{1,\cdots, n)\}$ according to their distance to $x$, $d_{pro}$. That is, for all $j<m$, $d_{pro}(x,x_{\pi_j(x)}) \leq d_{pro}(x,x_{\pi_{j+1}(x)}). $ Let $\{C_1,\cdots,C_N\}$ be an $\varepsilon$-cover of $\mathcal {X}$ of cardinality $N=\mathcal {N}(\varepsilon, \mathcal {X},d)$.  Eq.(19.3) in \cite{Shalev-Shwartz:2014:UML:2621980} implies that
	\begin{equation}\label{proof11}
		\begin{split}
			& E_{D \sim \mathcal {D}^n,(x,y)\sim \mathcal {D}} \Big(  P(y_i \neq h_{knn_i}^D(x))\Big)\\
			\leq  &  E_{D \sim \mathcal {D}^n} \Big( \sum\limits_{j:|C_j \cap D| < k} P(C_j)  \Big) \\
			+ & \max\limits_z P_{D \sim \mathcal {D}^n,(x,y)\sim \mathcal {D}}  \Big( y_i \neq h_{knn_i}^D(x) |
			\forall z \in [k], \\
			&d_{pro}(x,x_{\pi_z(x)}) \leq ||\hat{V}||_F\varepsilon  \Big)
		\end{split}
	\end{equation}
	Following the proof of Theorem 19.5 in \cite{Shalev-Shwartz:2014:UML:2621980}, the first term of the right side of Eq.(\ref{proof11}) is bounded by $\frac{2Nk}{en}$. The second term of the right side of Eq.(\ref{proof11}) is bounded by $(1+\sqrt{8/k})P(b_i^*(x)\neq y_i) +3L||\hat{V}||_F\varepsilon$. By setting $\varepsilon=2n^{-\frac{1}{\mathbb{D}+1}}$ and combining Theorem~3 , we get
	\begin{equation}\label{proof10}
		\begin{split}
			&E_{D \sim \mathcal {D}^n,(x,y)\sim \mathcal {D}} \Big( P(y_i \neq h_{knn_i}^D(x))\Big)\\
			\leq & (1+\sqrt{8/k})P(b_i^*(x)\neq y_i) \\
			+ &\frac{6L||\hat{V}||_F +  k}{n^{1/(\mathbb{D}+1)}}\\
		\end{split}
	\end{equation}
	We apply Eq.(\ref{proof10}) for each label and take the sum to derive the result.
\end{proof}

\footnotesize

\bibliographystyle{IEEEtranN}

\end{document}